\documentclass[11pt,a4paper]{article}

\usepackage[hyperref]{acl2020}
\usepackage{times}
\usepackage{latexsym}
\usepackage{times}
\usepackage{amsmath}
\usepackage{graphicx}
\usepackage{ dsfont }
\usepackage{amssymb}
\usepackage{tikz}
\usetikzlibrary{bayesnet}
\usepackage{url}
\usepackage{hyperref}
\usepackage{subcaption}
\usepackage{natbib}

\usepackage{float}
\newfloat{algorithm}{t}{lop}

\usepackage{algpseudocode}
\usepackage{microtype}

\aclfinalcopy % Uncomment this line for the final submission
\title{Learning Manner of Execution from Partial Corrections}

\author{Mattias Appelgren \\
 University of Edinburgh \\
  \texttt{mattias.appelgren@ed.ac.uk} \\\And
  Alex Lascarides \\
 University of Edinburgh \\
  \texttt{alex@inf.ed.ac.uk} \\}

\date{}

\begin{document}
\maketitle
\begin{abstract}

Some actions must be executed in different ways depending on the context. For example, wiping away marker requires vigorous force while wiping away almonds requires more gentle force.  In this paper we provide a model where an agent learns which manner of action execution to use in which context, drawing on evidence from trial and error and verbal corrections when it makes a mistake (e.g., ``no, gently'').  The learner starts out with a domain model that lacks the concepts denoted by the words in the teacher's feedback; both the words describing the context (e.g., marker) and the adverbs like ``gently''.  We show that through the the semantics of coherence, our agent can perform the symbol grounding that's necessary for exploiting the teacher's feedback so as to solve its domain-level planning problem: to perform its actions in the current context in the right way. 

\end{abstract}

\section{Introduction}

Recent work aims to teach robotic agents the manner in which to perform its actions, because some actions are similar in nature but need to be performed in different ways in different contexts. When wiping away a substance from a surface, the goal of the wiping action and the bulk of the movement will be the same across different substances, but there will also be required differences. For example, wiping away marker from a table requires pressing hard and moving quickly back and forth, while wiping almonds requires similar trajectories of movement but executed more slowly and gently (to avoid knocking the almonds everywhere).  

This paper describes and evaluates a model that tackles the task of learning which manner to execute in which context via 
evidence gathered in an Interactive Task Learning paradigm (ITL, \citealt{Laird2017InteractiveTL, ITLbook}): i.e., the agent learns from its actions and from contextually relevant teacher guidance.  Specifically, when the agent makes a mistake a teacher provides corrective feedback (e.g., ``no, do it gently'').  As this extended embodied interaction proceeds, our model supports online incremental learning of two things.  The first is the symbol grounding that's necessary for inferring the teacher's communicative intent: i.e., learning to map the words the teacher uses to the embodied environment, in particular adverbs like ``gently'.  The second is a domain-level policy for choosing the appropriate manner for performing the action in the current state.  The task the agent faces in our experiments is analogous to learning how to wipe different substances, selecting a manner in which to perform the action that's dependent on that substance.  We demonstrate that exploiting theories of discourse coherence \cite{asher2003logics}, and in particular the semantics of coherent corrections, enhances the performance of the model.

\section{Related Work}

\citet{Hristov2020LearningFD} train a model which disentangles a behaviour into different factors such as the speed or energy used, creating a vector space representing different manners for performing an action (fast vs. slow, vigorously vs. gently). Given this feature space, they train a model to generate an instance of the desired behaviour.  But the model relies on batch training offline, using a large dataset of labelled movements.  In contrast, there will be scenarios where the user wants to introduce a novel manner for performing an action {\em after} the agent is deployed, perhaps because a novel substance that needs to be wiped gets introduced into the domain after deployment.  In such a case, the user will want to teach the robot about the new substance, the new manner and when to use it.  Accordingly, in this paper we use 
Interactive Task Learning (ITL, \citealt{Laird2017InteractiveTL, ITLbook}) to teach an agent various ways to perform an action and which manner to use in which context. In ITL, as the extended embodied dialogue with a teacher proceeds, the learner must update its model of the domain, of embodied language understanding and its policies whenever new evidence is observed in the interaction (ie., our aim is to support learning incrementally and online). We focus on correction as the main mode of interaction---a useful teaching strategy because corrections can articulate what's wrong with the learner's latest action.

Corrective feedback introduces a novel challenge for embodied agents that need to acquire a model of symbol grounding---ie, a mapping from symbols to their denotations in the environment \cite{Harnad1990TheSG, Matuszek2018GroundedLL}. Symbol grounding in ITL has mainly been studied in the context of instruction giving and description, and so the symbols in the utterance denote something in the current scene (cf \citet{DBLP:conf/aaai/TellexKDWBTR11, Matuszek2018GroundedLL}). With corrections, some terms lack a denotation in the current scene \citep{Appelgren2020}: e.g., ``no, do it gently'' implies the manner just executed was {\em not} gentle. Thus our agent must infer from each utterance which concepts are denoted in the current scene, which concepts aren't, and how to use this evidence to update its knowledge of the world. We will utilise theories of coherence \citep{asher2003logics} to enable those inferences and updates to take place. 

Previous work on corrections has mainly focused either on demonstrating corrections physically (in Learning from Demonstration scenarios) or only as very simple ``yes''/``no'' feedback \citep{DBLP:conf/atal/NicolescuM03, DBLP:conf/kcap/KnoxS09}. Our interest, following \citet{Appelgren2020}, is to extend correction capabilities to more complex natural language utterances: the utterance doesn't simply stipulate that the agent's latest action was wrong, but also stipulates what should have been done instead. In this paper, we extend this prior work in two ways.  First, we tackle a new planning problem that involves not just grounding colours but also grounding adverbs. Second, we cover another natural way of correcting, which would occur frequently in human-robot interaction: namely `partial' corrections, in which the teacher does not articulate the feature(s) of the current context that demand the different manner that the teacher describes (e.g., ``no, gently''), and so the learner must infer those features to correct its mistake in future.

\section{Dialogue Strategy}
\label{sec:gently-dialogue}

Coherence-based theories of discourse model how the meaning of the current utterance is constrained by the way in which it is semantically connected to its context \citep{asher2003logics, hobbs:1979}.
In this paper, we will exploit the semantics of the coherence relation {\em correction} to inform our agent's inferences about the meaning of the teacher's utterance and the resulting update to its knowledge. 

We'll illustrate the approach with the running example of wiping almonds or marker. We assume there is a required way to wipe each substance: gently and slowly for almonds; quickly and firmly for marker. The agent starts out unaware of the adverbs {\em slowly}, {\em gently} etc, and also lacks their denoted concepts in its domain model---it can observe speed and force, but doesn't know how the continuous spectra of values partition into the commonsense concepts that humans talk about.  Nor does the agent's domain model include almonds or marker, and (so) the agent cannot ground the terms ``almonds'' and ``marker''.  Likewise, the agent doesn't know the manner in which it should wipe almonds or marker.  With this in mind, in this paper we assume the teacher can say three types of things (allowing terms to be replaced as appropriate):
\begin{enumerate}
    \item yes
    \item no, wipe almonds gently and slowly
    \item no, do it quickly and firmly
\end{enumerate}
We will assume that all the teacher's utterances are true and follow a coherent strategy.
The question then becomes what the agent can learn from each utterance.

Let us examine a few scenarios. First, the agent has wiped almonds and has done it gently and slowly (although the agent may not know what it's just wiped nor that its manner of execution can be described this way). In this situation it would seem incorrect to say ``no, wipe almonds gently and slowly'', ``no, wipe marker quickly and firmly'' or ``no, do it quickly and firmly". In all these cases ``no'' signals the utterance corrects the action and so the action exhibits the wrong manner, but this is false.  But more fundamentally, one cannot identify a coherent and satisfiable relationship between the resolved meaning of ``no'' (in this context, do not wipe the almonds in the manner you did) and the matrix clause: the first is contradictory; the second is a non-sequitur; and the third, with the pronoun resolved to ``almonds" (as demanded by coherence), is false. The only valid response to the agent's (correct) action is ``yes'' or some equivalent. 

Now assume the agent wiped almonds quickly and firmly. Then the teacher shouldn't say ``yes'' (because its resolved meaning is false), and ``no, wipe marker quickly and firmly'' remains a non-sequitur.  But it would be coherent and true to say ``no, wipe almonds gently'', ``no, wipe almonds slowly'', ``no, wipe almonds slowly and gently'', or ``no, do it slowly and gently''. 

% Can we explain these intuitions and make any generalisation which is helpful to our agent? \citet{lascarides:stone:2009} propose a semantics for coherent corrections, stating that they are coherent if they are {\em relevant} and {\em deny} some part of the corrected move. When we say ``no, wipe almonds gently and slowly'' when the agent has just wiped almonds gently and slowly we aren't denying anything, and so it seems incoherent. When the the teacher says ``no, wipe marker quickly and firmly'' when the agent has just wiped almonds the correction is irrelevant because marker has nothing to do with the action at hand, it is {\em irrelevant}. 

More generally, an interlocutor infers from a coherent and true correction that it is relevant (in our case, it's related to the agent's latest action) and denies some aspect of it. So if the agent is told ``no, wipe almonds gently and slowly'' it knows that it just wiped some almonds (even if it didn't believe this before), and furthermore did not do it gently and/or slowly. The agent also knows when it wipes almonds it must do so gently and slowly. It might have believed it was doing so, in which case it would now have to revise its beliefs about what slowly and gently denote. In the remainder of this paper, we will show how the agent can use these facts to update its knowledge and learn.

\section{The Task: Learning which Manner of Execution in which Context}
\label{sec:gently-task}

\begin{figure}
    \centering
    \includegraphics[width=0.7\linewidth]{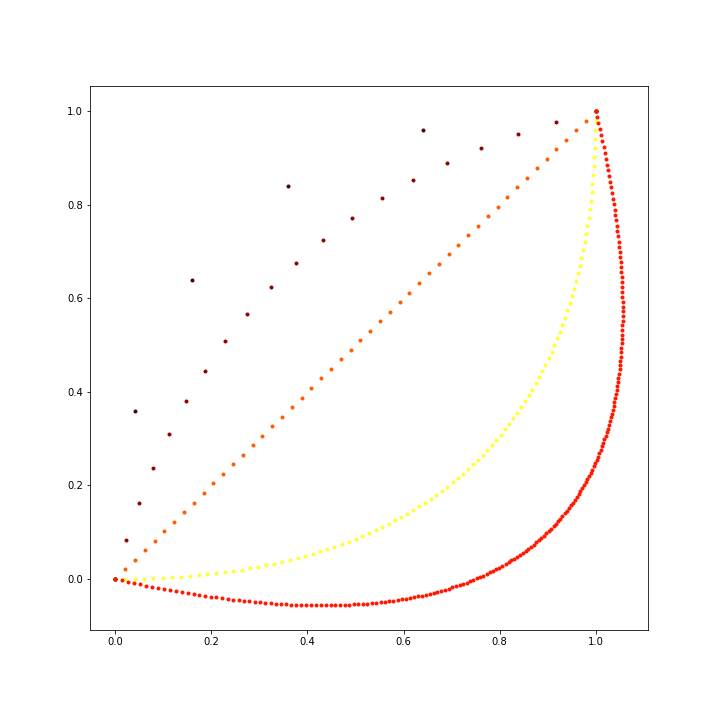}
    \caption{Five examples of different curves representing the manner in which the agent performed an action. Darker red colours represent lower energy, lighter yellow colours represent high energy. Few dots represent slow actions while more dots represent fast actions. }
    \label{fig:gently-curves}
\end{figure}

\begin{figure}
    \centering
    \includegraphics[width=0.7\linewidth]{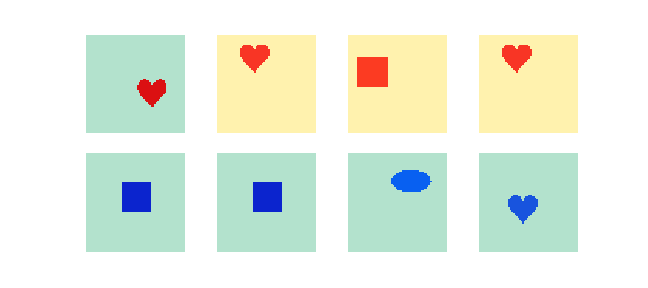}
    \caption{Examples of colours and shape situations as presented to the agent. }
    \label{fig:gently-situations}
\end{figure}

The task our agent will tackle is a simplified version of the wiping problem we have described so far. A generalisation of the task is that there is an action that needs to be performed, the agent can select to perform the action in different ways, and there are certain contexts where it is required to perform the action in particular ways. We create an abstract version of the wiping task which removes some of the noise of real-world wiping. Specifically, the context will be represented by a coloured shape, as shown in Figure \ref{fig:gently-situations}. For selecting the manner in which to perform the action, following \citet{Hristov2020LearningFD} we will assume we have a disentangled space where each dimension represents a different dimension of manner. We use three dimensions representing speed, energy, and direction. We represent these visually using Bezier curves where the speed changes the number of dots used to draw the curve, the energy determines the colour, and direction determines where the midpoint of the curve lies; see Figure \ref{fig:gently-curves}. 

The teacher knows all the constraints on the manner of execution, which are dependent on the particular shapes and/or colour of the object the action is performed on.  In contrast, at the start of the learning process, the agent can recognise shapes, and it knows rules exist that require a particular behaviour in particular situations. The agent also knows there are concepts describing manner and knows which dimension of behaviour space each concept is on, but does not know which region of the behaviour space it is on.  The agent's domain model likewise lacks all colour categories---although it can observe RGB values---and the agent's natural language vocabulary doesn't include the words denoting colour.

The agent is shown a coloured shape and must select a point in behaviour space to perform an abstract `action'. The teacher is shown the coloured shape and the resulting Bezier curve. The teacher then gives feedback by correcting the agent if the agent's action on the object violates one of the constraints.

\section{System Description}

The agent needs to be able to do two things. First, it needs to be able to select an action to perform. Selecting an action in our task is equivalent to selecting a point in {\em behaviour space}. The selection must be done based on the current {\em context} and the agent's current beliefs about the constraints on manner of execution. The second thing it needs to do is update its knowledge given the available evidence. It will do so based on the analysis of coherent correction, presented in Section \ref{sec:gently-dialogue}. 

\subsection{Learning from Correction}

The agent has just observed situation $s$, it decided to perform action $a$ in a particular manner which was determined by selecting a point $p$ in behaviour space, i.e. selecting a speed, energy, and direction of the action. In this situation the teacher makes one of the following three types of utterances:
\begin{quote}
    $u_1 = $ ``no, when you see red squares do it gently and slowly''\\
    $u_2 = $ ``no, do it quickly''\\
    $u_3 = $ ``yes''
\end{quote}

The agent needs to learn three types of knowledge to act correctly: 
\begin{enumerate}
    \item rules describing in what situations particular manners of execution are required
    \item a mapping between observable features and the concepts the teacher uses to describe situations
    \item how to generate the correct actions, given particular adverbs
\end{enumerate}

The agent will infer which rules exist based on the teacher's guidance. Each rule is represented as a {\em body} which can be either shape, colour, or both, and a {\em head} which will be a single adverb: for example, $red(s) \rightarrow gently$. To capture the agent's belief we keep a probability distribution $P(gently|red(s))$, which has a high probability if the agent believes the rule constrains the problem and a low probability if it thinks it does not (see \S\ref{sec:learn-rules}).

The second problem amounts to grounding \citep{Matuszek2018GroundedLL}: i.e., associating the visual features of a situation $F(s)$ with whether or not an object can be described as a concept, e.g. red. We represent this with $P(red(s)|F(s))$ (see \S\ref{sec:colour-grounding}). 

Finally, given it has selected to perform an action in a particular manner, say gently and slowly, the agent must generate a particular point $p$ in behaviour space. That is, it must select a value for energy, speed, and direction. To do this we will learn a normal distribution for each concept which will generate points in a particular region of one of these dimensions. The agent knows on which dimension a concept falls so must simply find the region within the dimension. We keep a probability distribution $P(p|gently) = N(p: \mu_g, \sigma^2_g)$ (see \S\ref{sec:learn-behaviour}).

\subsubsection{Learning Rules}
\label{sec:learn-rules}

To begin with we look at how the agent learns the rules from utterances $u_1$, $u_2$, and $u_3$. Starting with $u_1$ = ``when you see a red square do it gently and slowly'' this means there are two rules under our definition of rules: $r_1 = \text{red}(s) \wedge \text{square}(s) \rightarrow \text{gently}(a)$ and $r_2 = \text{red}(s) \wedge \text{square}(s) \rightarrow \text{slowly}(a)$.\footnote{We could have represented this as $\mathit{red}(s) \wedge \mathit{square}(s) \rightarrow \mathit{gently}(a) \wedge \mathit{slowly}(a)$ but it makes more sense in the probability model to keep them separate.} Given that the teacher utterances are always true, the agent's belief should be: $P(gently|red(s), square(s))=1$ and $P(slowly|red(s), square(s))=1$. 

Things get more difficult when the teacher says $u_2$ = ``no, do it quickly''. $u_2$ states a behaviour which needs to be performed (``gently''), but not the features of the context that constrain the required behaviour to quickly. However, we know from our analysis of coherence that it would only be said if it was relevant. In other words, the current context would have to match the body of the rule in question. Assuming the agent knows that the current object is a red square, then there are three possible rules:
\begin{enumerate}
    \item $red(s) \wedge square(s) \rightarrow quickly(a)$
    \item  $red(s) \rightarrow quickly(a)$
    \item  $square(s) \rightarrow quickly(a)$
\end{enumerate}
It is impossible from a single correction to know which of these is the true rule. The agent could ask a question to clarify, but for the purpose of this paper we'll assume the agent will try to learn which rule is correct from subsequent evidence.  Thus the agent must estimate its belief in each of the rules: that is, estimate the beliefs $P(quickly|red(s), square(s))$, $P(quickly|red(s))$, and $P(quickly|square(s))$. The belief is represented by a probability which we parameterise with a Beta distribution with parameters  $\alpha$ and $\beta$ so a rule's belief is captured as follows:
\begin{gather}
    P(\textit{quickly}(a)|\textit{red}(s), \textit{square}(s)) = \\ 
    \nonumber Beta(\alpha, \beta) = \frac{\alpha}{\alpha + \beta}
\end{gather}
$\alpha$ represents positive evidence for the rule while $\beta$ represents negative evidence. $u_2$ gives positive evidence for the three rules described above meaning we would add $1$ to $\alpha$. However, if the agent is told ``yes'' when it performed the action slowly and there was a red object then $red(s) \rightarrow quickly(a)$ is probably not true. The agent would therefore add $1$ to the $\beta$ parameter. Over time, the agent strengthens its belief about which rules truly are part of the problem. However, we must also take into account the agent's uncertainty about the true world state. To do this, instead of adding $1$ we use the probability that the state is, e.g., red and square for rule 1, or just red for rule 2. I.e. we add $P(red(s), square(s)|F(s))$ to $\alpha$.

\subsubsection{Learning Colour Grounding}
\label{sec:colour-grounding}

To learn to ground concepts, the agent needs to learn the probability $P(red(s)|F(s))$. We treat this probability as binary between being red or not red, rather than having a multimonial distribution with possible values red, green, blue, etc. We do this because our agent starts out without the knowledge of which colour categories exist, and so adding a new (and unforeseen) colour category amounts to adding a new binary classifier. Further, certain colour terms aren't mutually exclusive: a maroon object is also red. 

To learn the probability distribution the agent must gather data $D_{c} = \{(w_1, F(s_1)), ...(w_m, F(s_m))\}$ where $c$ is a concept and $w_i$ is a weight indicating how likely the agent believes it is that $s_i$ is $c$.  It does so by using the observation stated in Section \ref{sec:gently-dialogue}, namely that when the teacher says $u_1$ the utterance must be relevant. It is only relevant if the {\em context} described in the utterance is true in the current situation: for $u_1$ the object must be a red square, and so the learner adds $(1, F(s))$ to $D_{red}$ and $D_{square}$. This means that every utterance which states a context tells the agent more about how to ground an object. However, the agent cannot learn anything about grounding from utterances $u_2$ and $u_3$ because neither of them provides any information about the concepts which might apply to the current situation. 

We use Bayes rule to estimate the probability $P(red(s)|F(s))$:
\begin{equation}
P(red(s)|F(s)) = \frac{P(F(s)|red(s))P(red(s))}{P(F(s))}
\end{equation} For the prior $P(red(s))$ we choose $0.2$. For $P(F(s)|red(s))$ we use a weighted Kernel Density Model (wKDE). The probability is defined by the equation
\begin{equation} \label{eq:wkde}
P(F(s)|\text{red}(s)) =\frac{1}{\sum\limits_{i=1}^{m}w_{i}}\sum_{i=1}^{m}w_{i}\cdot\varphi(F(s)-F(s_{i})) 
\end{equation}
with data taken from $D_{red}$ and with kernel function $\varphi$. We use a Gaussian distribution as the kernel.  

\subsubsection{Learning to Generate Behaviour}
\label{sec:learn-behaviour}

In this work we assume we have a disentangled space for generating the manner in which an action is performed. Thus manner is represented by points in ``behaviour space'' and the agent must figure out which points correspond to each adverb. We are going to assume that a range of different points could satisfy the same adverb and we will use a Gaussian distribution to model which points should be generated by an adverb, e.g. $P(p|gently)=N(p: \mu_{g}, \sigma_g)$.

The data for learning $\mu_g$ and $\sigma_g$ must be gained from interaction. We know from Section \ref{sec:gently-dialogue} that when the teacher says ``no, do it gently'' or ``no, when you see red squares do it gently'', then the agent did {\em not} do it gently. I.e. it is a negative exemplar of doing it gently. Thus the corrections give us a set of data, $X_-$, of negative exemplars of gently.  But normally generative distributions, of which a normal distribution is an example, are trained on positive exemplars. The agent also benefits from positive data to train on from these interactions, specifically when the teacher says ``yes''. In this case, the agent knows that all constraints were adhered to, otherwise it would be corrected. Therefore, if the agent is confident that (i) there's a rule $r$ with body $b$ and head $h$; and (ii) $b$ applies to the current situation, then it knows it must have performed the action according to $h$. So, for instance, if the agent is confident that the rule $red(s)\rightarrow gently$ is true and also confident that the shape in $s$ is red, then the agent can conclude that it has performed the action gently. From these situations we can create a set of positive data $X_+$. The agent decides to add a datapoint to $X_+$ when $P(red(s), r) > 0.7$. 

To learn the parameters $\mu_g$ and $\sigma_g$, we create a KDE model of the positive data $X_+$ and negative data $X_-$, resulting in a distribution that captures which region is more or less likely to be the correct region for ``gentle'' actions. These models are then used to classify points placed regularly on the interval between highest and lowest on the behaviour dimension (for ``gently'' that's the energy dimension). The parameters are estimated by computing the mean and variance of the points with probability bigger than $0.5$ of being good. This allows the agent to exclude regions of the space when it receives negative data, which can speed up the rate at which it converges to an accurate model of behaviour generation. 

% \begin{algorithm}
% \begin{algorithmic}
%     \Function{UpdateBehaviour}{$X_+$, $X_-$} 
    
%     \State $P_+ =$ \Call{KDE}{$X_+$}
%     \State $P_- =$ \Call{KDE}{$X_-$}
%     \State $pts =$ \Call{linspace}{$min$, $max$, $50$}
%     \State ${points}_+ = P_+(pts)$
%     \State ${points}_- = P_-(pts)$
%     \State $p\_points = [p_+/(p_+ + p_-)\ \texttt{for}\ p_+,\ p_-\ \texttt{in}\ {points}_+, {points}_-]$
%     \State ${good\_points} = [p\ \texttt{for}\ p\ \texttt{in}\  {p\_points}\ \texttt{if}\ p > 0.5]$
%     \State $\mu, \sigma =$ \Call{mean}{$good\_points$}, \Call{var}{$good\_points$}
%     \State \Return $\mu, \sigma$
    
%     \EndFunction

% \end{algorithmic}
% \caption{Algorithm for updating the parameters of Gaussian Distributions for generating behaviour. It uses the known data to generate a set of ``good'' points taken from the relevant dimension in behaviour space. These good points are used to update the parameters of the Gaussian using a straightforward Bayesian update. }
% \label{alg:update-behaviour}
% \end{algorithm}

\subsection{Action Selection}
\label{sec:action-selection}

When faced with the current object, the agent must choose the action it will perform (i.e., a point $p$ on the action space).  Its choice is made dependent on its current beliefs about the constraints on manner (\S\ref{sec:learn-rules}), the mapping of concepts that are expressed in those rules to the current visual features (\S\ref{sec:colour-grounding}), and its beliefs about how to generate particular manners like gently (\S\ref{sec:learn-behaviour}). 

Those beliefs, and the inferences they afford to beliefs about which behaviour points $p$ are constraint-compliant in the current situation $s$, are captured in a Bayes Net: its graphical component expands dynamically as the learner discovers unforeseen concepts through the interaction, and its parameters get incrementally updated as the interaction proceeds, as we've already discussed.  

When the teacher mentions a new colour term $C$ (via a full correction of the form $u_1$, which also features shape $S$ and manner $M$), it triggers the introduction of a new Boolean chance node $C$, a dependency from the observable visual features (RGB values) $F(s)$ to $C$, a dependency from $C$ to $M$, and (if it doesn't already exist) a dependency from the (existing) observable shape node $S$ to the manner node $M$.  If $M$ is also new, then likewise this triggers a new chance node $M$, which in turn triggers a dependency to the appropriate dimension of behaviour (e.g., speed for ``quickly'' and ``slowly'').  The parameters on all these nodes (existing and new ones) are then computed according to our earlier discussion (see \S\ref{sec:colour-grounding} for $C$, \S~\ref{sec:learn-rules} for node $M$ and \S\ref{sec:learn-behaviour} for estimating $P(p|M)$).

If a novel manner $M$ is introduced in a partial correction $u_3$ (e.g., ``no, gently''), then the newly introduced chance node $M$ has as parents the shape $S$  that's observed in the current state, and all the existing colour nodes. Again, its parameters get computed as described in \S\ref{sec:learn-rules}, and its connected to the appropriate behaviour dimensions as described in \S\ref{sec:learn-behaviour}.  These multiple parents for $M$ reflect the fact that the partial correction is consistent with several possible rules. 

\begin{figure}
    \centering
    \includegraphics[width=\linewidth]{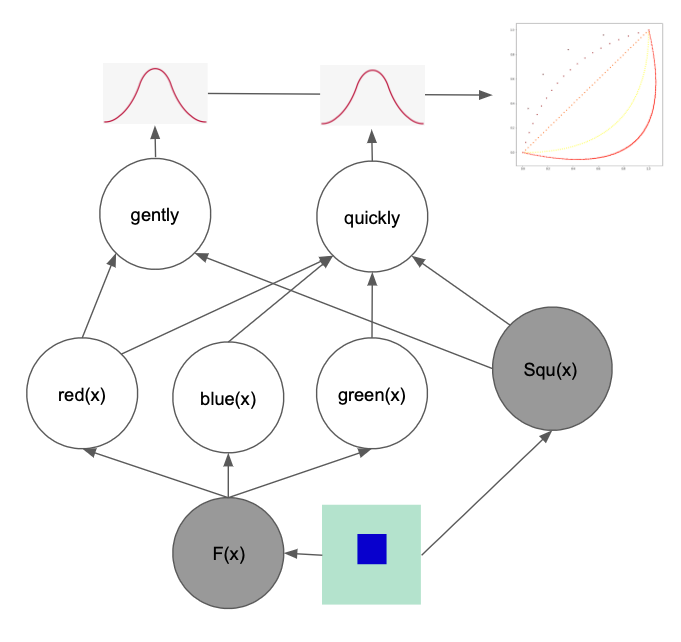}
    \caption{The graphical component of the agent's Bayes Net when in a prior move the teacher said ``when you see red squares do it gently'', likewise in prior moves the teacher has mentioned ``green'' and ``blue'', and the teacher now says ``no, quickly''. Grey nodes are observed and white nodes are latent.}
    \label{fig:gently-model}
\end{figure}

Figure~\ref{fig:gently-model} illustrates the graphical component of a specific example Bayes Net that the agent has constructed via the interaction so far.  The agent's decision making then consists of two stages.  The first stage is to estimate the set of adverbs that describe the manner in which the actions should be executed in the current situation.  The second stage is to then generate a specific point in behaviour space that's accurately described by those adverbs.

Assuming
we know of the four concepts gently, firmly, slowly, quickly and assuming that adverbs on the same dimension denote different portions of it (unlike colour categories), there are 9 different options for what behaviour to do: each combination of gently/firmly and quickly/slowly makes 4, plus each individual concept separately adds 4, and finally we could have no preference, i.e. no adverbs. 
The decision problem is then to select which of these 9 options is the most likely. If we have selected some adverbs, say gently and quickly, then the probability is calculated by setting those adverbs as positive and the rest as negative in the model:
\begin{align}
    P&(gently, quickly|F(s)) = 
     \\\nonumber P&(gently, quickly,\neg slowly, \neg firmly|F(s))
\end{align}
We calculate this probability by using Variable Elimination on the current Bayes Net; e.g., Figure \ref{fig:gently-model}. 

% The model is built dynamically based on the agent's knowledge of the rules and colours that exist. For example, if the agent has been told ``when you see red blocks do it gently'' we add to the model a factor for the rule: $P(gently|red(s), block(s))$ as well as factors for recognising red blocks: $P(red(s)|F(s))$ and $P(square(s)|F(s))$. When there are several possible rules that constrain the action to being performed gently, then we add nodes as appropriate for each of those rules. 

Inference simply finds the most likely out of the 9 different options, resulting in some number of behaviour concepts that need to be generated from. Generation is done simply by using the model for generating a point in behaviour space based on an adverb (see \S\ref{sec:learn-behaviour}). Each adverb is associated with one dimension of behaviour space and so each adverb generates a point on their associated dimension. A point in any remaining dimension is sampled from a uniform distribution.

\section{Experiments}

We aim to show that an agent can effectively learn new concepts and constraints through an interaction with a teacher where the main mode of interaction is correction.  In particular, we aim to show that learning from more contentful corrections is more data efficient than just learning from ``no''. Further, we provide two ablation studies to show that its important to learn when the teacher says ``yes'' and its important to learn from the negative datapoints the agent gets for the adverbs when the teacher corrects the agent. 

Each experiment is made up of five trials consisting of 100 different situations. The agent observes each situation in turn, observing what shape is depicted and its RGB value. Given these observations, the agent must select a point in behaviour space which is used to generate a behaviour curve, as described in Section~\ref{sec:action-selection}. The teacher observes the generated behaviour and gives feedback, either correcting it or saying ``yes''. To ensure that the agent mainly faces scenarios where it has to make an interesting decision, the scenarios are generated so that 90\% of them contain contexts which are constrained by the rules, with the remaining 10\% taken randomly from the set of all possible states.

We run two experiments. In the first the teacher always expresses a full rule using expressions like $u_1$ while in the second experiment the teacher expresses half the rules only partially, using the expression $u_2$. In the second experiment the constraints are chosen so that the colour in a rule that triggers a partial correction is also mentioned somewhere in a full correction: otherwise, the agent would be unable to learn the rule because it would be unaware of the colour category that expresses the rule.
To evaluate the agents learning we measure regret, which we define to be the number of scenarios in which the agent does not perform a correct action---i.e., it is the number of times it gets corrected. 

In each experiment we test five different agent strategies. The first is a Random baseline which simply selects a point in behaviour space uniformly at random and does not attempt to learn at all. Second, the ``Just No'' agent learns only from the teacher saying ``yes'' or ``no''. This agent creates a basic rule whose body is single datapoint classifier for the current state, and it head says that everything which that classifier catches should be done either in the same way for ``yes'' or not in the same way for ``no'' (so it fails to recognise that red denotes a wide range of RGB values, and adverbs an extended range on the Bezier curves). Third, we have our full agent as described in this paper, which learns from the entirety of the teacher's utterances. Finally, we have two ablation baselines. The first makes no update when the teacher says ``yes''---it does not learn from assent. The second does not update the behaviour concept models with negative data points $X_-$, it only learns from situations where it can add a positive data point to $X_+$. These models are meant to show the value of using coherence to learn not just from the correction but from assent, and to show that our method for learning from the negative data points, as supported by the semantics of coherent corrections, speeds up learning. 

\begin{figure*}
    \centering
    \begin{subfigure}[t]{0.45\textwidth}
    \centering
        \includegraphics[width=0.8\linewidth]{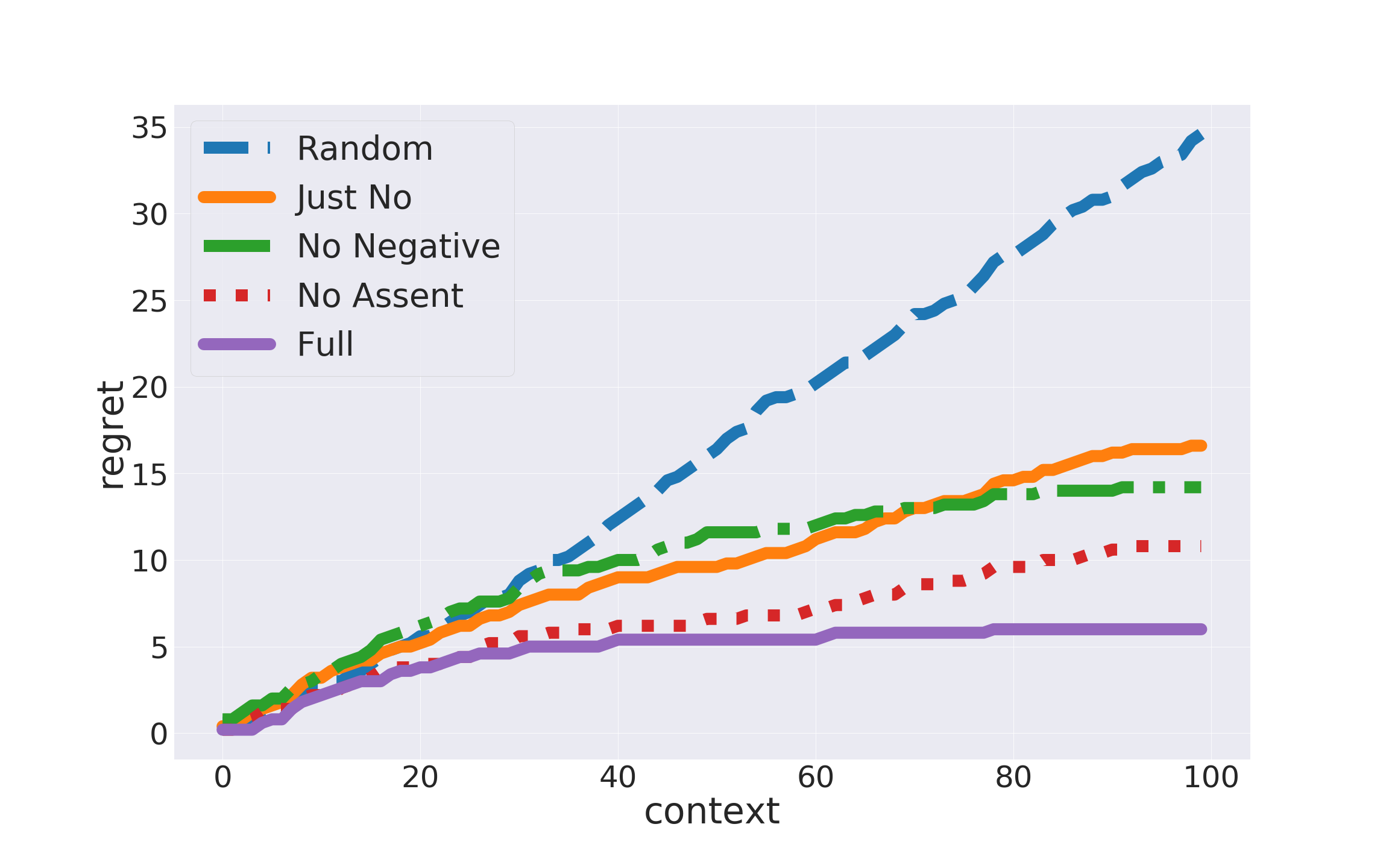}
        \caption{The teacher expresses all constraints fully.}
    \end{subfigure}
    \begin{subfigure}[t]{0.45\textwidth}
    \centering
    \includegraphics[width=0.8\linewidth]{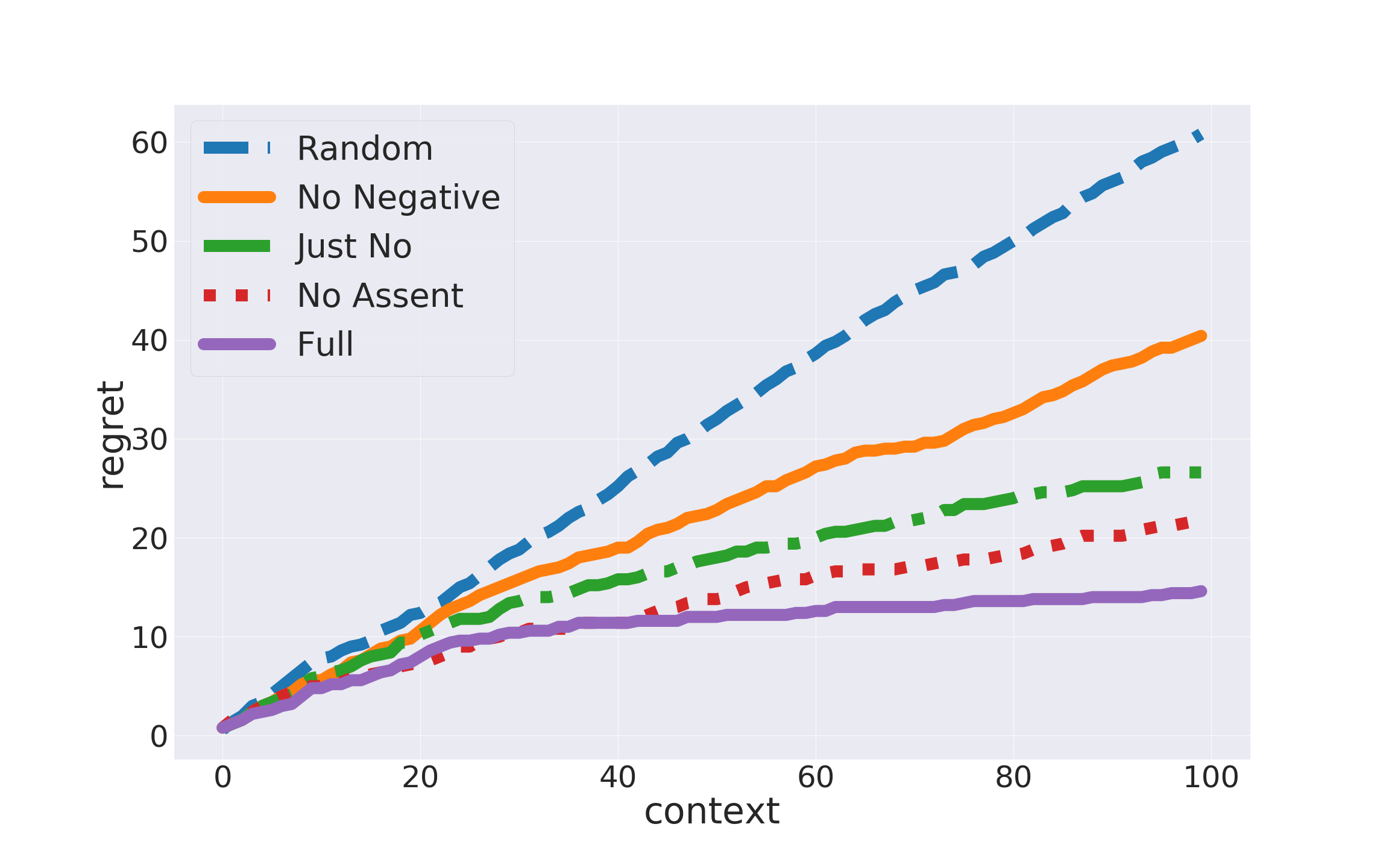}
    \caption{The teacher expresses half of the constraints fully and half partially.}
    \end{subfigure}
    \caption{Results for two experiments comparing our Full agent against a Random baseline, an agent that learns just from the fact that the teacher says yes and no, and two ablations. The No Assent agent does not update when the teacher says ``yes'' and the No Negative agent does not use negative exemplars to update its models for generating behaviours. }
    \label{fig:gently-experiments}
\end{figure*}

\begin{table*}
    \centering
    \begin{tabular}{c|c c c c c}
       Dataset & Random & No Negative & Just No & No Assent & Full  \\\hline
       Partial Corrections  & $31\pm 18 $ & $22\pm 13 $ & $16\pm 8 $ & $13\pm 6$ & $11\pm 4$  \\
       Fully Expressed  & $16\pm 11 $ & $10\pm 7 $ & $10\pm 5 $ & $7\pm 3$ & $5\pm 2$  \\
      
    \end{tabular}
    \caption{Mean and standard deviation of terminal regret for our different agents on two different datasets. }
    \label{tab:summary-gently}
\end{table*}

\subsection{Results}

Figure~\ref{fig:gently-experiments} shows the results of the experiments with the mean terminal regret summarised in Table \ref{tab:summary-gently}. The ``Just No'' agent manages to learn about the task in both experiments, acting much better than randomly ($t=9.25$ $p=6.82e^{-6}$). However, our full agent clearly outperforms the ``Just No'' agent ($t=7.91$ $p=2.43e^{-5}$). This is especially true given the regret curve in Figure~\ref{fig:gently-experiments}. The full agent essentially makes no mistakes during the final scenarios while the ``Just No'' agent still performs errors. This supports our hypothesis that learning from contentful corrections will be more data efficient than learning just from ``yes''/``no'' feedback. 

For the ablation study in which the agent ignores assent, in both experiments this hurts performance ($t=4.00$ $p=3.09e^{-3}$). So exploiting assent improves data efficiency.  Correction gives the agent {\em positive} exemplars of which {\em context} it is in, but mostly gives {\em negative} exemplars of the {\em required manner}. Assent is a perfect complement to this since it provides positive exemplars of the required manner but relies on the knowledge gained from the correction in order to do so. 

In the second ablation the agent does not update its models of behaviour concepts with negative data points. Comparing the curves for this agent in Figures~\ref{fig:gently-experiments}a and~\ref{fig:gently-experiments}b suggests that this significantly hampers performance, especially with the teacher expressing partial corrections.  Table~\ref{tab:summary-gently} also shows that ignoring the negative data hurts performance compared to the full agent ($t=3.74$ $p=4.63*10^{-3}$). It's unclear why the agent fails so spectacularly when it's denied the capacity to learn from negative exemplars. It's possible that it isn't confident enough to learn from enough positive exemplars alone when combined with uncertainty about the rules. 

\section{Conclusion}

In this paper we tackled the task of an agent learning the way it should perform an action depending on its context, with evidence coming from an extended interaction with a teacher.  The agent starts ignorant of how words denoting manner map to particular features of movement, unaware of domain-level concepts in which the constraints on manner in the planning problem are expressed, and ignorant of those constraints.  We have shown that learning from corrections that express a reason why an action is being corrected is more effective than learning from just saying ``yes'' or ``no''. The semantics of coherence guides the agent's learning process. 

We show that to learn efficiently, the agent can and should use both the fact that corrections tell it about positive exemplars of a context and negative exemplars for the required behaviour. When learning also from teacher assent, the agent can exploit coherence to acquire and learn from positive exemplars of required behaviour as well. This combination allows the agent to quickly learn to associate particular situations with particular required behaviours. In addition to this, we showed that the agent can still learn the required constraints on behaviour even when corrections express those constraints only partially.  It can do this by exploiting the semantics of coherent corrections combined with its (uncertain) interpretation of the scene to infer generalisations.

\section{Limitations}

We developed this work in the context of an ITL scenario because we're interested in tackling situations where an unforeseen factor or action gets added to the hypothesis space after the agent is deployed. We're very far from a general ITL agent, but the model presented here addresses a very natural type of teaching strategy, in which a teacher identifies the learner's mistakes and provides guidance on what the mistake was and/or how to do better. However, the effects of coherence were hard wired into the learning model.  To expand such techniques to more general discourse that will feature a wide range of coherence relations, we would need to augment this model with a generalised way of performing the pragmatic analysis of the teacher's communicative action, perhaps via prior (offline) learning of a discourse parser that estimates the coherence relation for the current utterance, combined with automated inference about the implications of that coherence relation.  This is a direction of future work to be considered. 

The second limitation of the current work is that we use an abstraction for both the underlying behaviour space and the range of situations (i.e., 2D coloured shapes, rather than 3D realistic images). The main architecture that we have proposed would still work with the more complex tasks but the exact models we use may not be appropriate. The main question which would arise is how to integrate our approach with more sophisticated learning models. The models we selected have fairly straight forwards methods for being updated incrementally, something which is vital when learning from dialogue (for the teacher wants observed differences in the learner's behaviour as the dialogue proceeds).  However,  \citeauthor{Hristov2020LearningFD}'s (\citeyear{Hristov2020LearningFD}) model and most state of the art vision models are trained using batch learning and are not designed to be incrementally updated nor for learning from just a few exemplars. 

\section{Ethics}

The work presented here does not directly harm or impact any people: we do not use any sensitive data and do not use human participants in our experiments.  

We believe this work could ultimately contribute to an increase in usability and accessibility to robotics in the future. If the goals of ITL can be realised, then anyone with the ability to interact with a robot would be able to teach it to execute tasks in their preferred way. However, we focus on language as the vector for teaching, which could exclude those without a voice. 
Further, while manipulating behaviour to the user's preferences can be beneficial, a malign user can create malign behaviours.  There is always a possibility of abuse or unintended use of such systems (like most systems that must respond to the user's needs). However, the scope of our system is limited and wouldn't be able to be exploited to perform any harmful action. For general ITL see a discussion of ethics in \citet{ITLbook}.

\bibliography{main}
\bibliographystyle{acl_natbib}

\end{document}